\documentclass{article}

\usepackage{PRIMEarxiv}

\usepackage[utf8]{inputenc} % allow utf-8 input
\usepackage[T1]{fontenc}    % use 8-bit T1 fonts
\usepackage{hyperref}       % hyperlinks
\usepackage{url}            % simple URL typesetting
\usepackage{booktabs}       % professional-quality tables
\usepackage{amsfonts}       % blackboard math symbols
\usepackage{nicefrac}       % compact symbols for 1/2, etc.
\usepackage{microtype}      % microtypography
\usepackage{lipsum}
\usepackage{fancyhdr}       % header
\usepackage{graphicx}       % graphics
\graphicspath{{media/}}     % organize your images and other figures under media/ folder
\usepackage{amsmath, amssymb, cases}
\usepackage{array} % Required for m column type
\usepackage{longtable}
\usepackage{amssymb} % for checkmarks
\usepackage{booktabs} % for better table formatting
\usepackage{color}
\usepackage{tabularx}
\usepackage{multirow}
\usepackage{geometry}
\newcommand{\redcircle}{{\color{red}\Large\textbullet}}
\newcommand{\orangecircle}{{\color{yellow}$\bigtriangleup$}}
\title{Large Language Models Meet Computer Vision; A Brief Survey
}

\author{
  Raby Hamadi \\
  Artificial Intelligence Department, Tahakom \\
   \\
  \texttt{rhamadi@tahakom.com} \\
}

\begin{document}
\maketitle

\begin{abstract}
Recently, the intersection of Large Language Models (LLMs) and Computer Vision (CV) has emerged as a pivotal area of research, driving significant advancements in the field of Artificial Intelligence (AI). As transformers have become the backbone of many state-of-the-art models in both Natural Language Processing (NLP) and CV, understanding their evolution and potential enhancements is crucial. This survey paper delves into the latest progressions in the domain of transformers and their subsequent successors, emphasizing their potential to revolutionize Vision Transformers (ViTs) and LLMs.  This survey also presents a comparative analysis, juxtaposing the performance metrics of several leading paid and open-source LLMs, shedding light on their strengths and areas of improvement as well as a literature review on how LLMs are being used to tackle vision related tasks. Furthermore, the survey presents a comprehensive collection of datasets employed to train LLMs, offering insights into the diverse data available to achieve high performance in various pre-training and downstream tasks of LLMs. The survey is concluded by highlighting open directions in the field, suggesting potential venues for future research and development. This survey aims to underscores the profound intersection of LLMs on CV, leading to a new era of integrated and advanced AI models.
\end{abstract}

% keywords can be removed

\section{Introduction}

The field of AI is witnessing a great fusion of CV and NLP. At the middle of this exciting combination are LLMs, which have come to be a key participant in AI. These models, that were first created to understand and produce human language, are actually expanding to encompass the interpretation of visual data. This shows how flexible and adaptable LLMs are, as they begin to combine the understanding of text with seeing and interpreting visual data. LLMs, along with CV are set to revolutionize the way machines understand and describe the visible world. This fusion of LLMs and CV is leading to an era where AI systems can see the world, understand it, and communicate about it nearly like humans do. By bringing together these technologies, we can make the way humans and machines interact smoother and more natural. For example, in security, this could mean systems that not only see something unusual but also explain it clearly, helping to deal with threats faster. In healthcare, combining LLMs and computer vision could lead to better diagnosis and treatment by joining visual information with a wide range of medical knowledge. In retail, AI could watch over inventory levels and use LLMs to make useful reports and predictions. In manufacturing, these technologies could improve how we check for product quality by spotting defects with computer vision and using LLMs to give detailed advice on fixing them. Looking ahead, the fusion of LLMs and computer vision is set to be a landmark in AI's progression, bringing us closer to creating highly autonomous digital assistants that can engage with the world comprehensively. This new era of AI envisions machines as proactive participants, interpreting and narrating the visual world around them.

A pivotal advancement in the fields of NLP and CV was the invention of the transformer architecture~\cite{46201}, which was crucial at managing long sequences in NLP tasks through the innovative attention mechanism~\cite{46201}. This mechanism allows the model to weigh the importance of different segments of input data. The breakthrough came when researchers stated that if transformers could interpret language as sequences of words, they might also process images as sequences of pixel patches. This insight gave rise to VT, a step forward in CV methodologies. In our survey, we aim to present an exhaustive exploration of LLMs as they venture into the realm of CV. We will delve into the nuances of transformer architectures and their successors, highlighting their role in shaping an integrated AI future. Our literature review will meticulously cover recent works and groundbreaking research that have charted the course for LLMs in visual applications. A comparative analysis will illuminate the strengths and potential enhancements of both paid and open-source models. Additionally, we will explore the diverse datasets that are used in the literature to train these models. The survey is then concluded by some of the challenges that face LLMs and some open research directions.

\section{Background}
\label{background_section}

\subsection{RNNs}
\label{rnn_subsection}
Recurrent Neural Networks (RNNs)~\cite{sherstinsky2020fundamentals} are a class of artificial neural networks designed for processing sequences of data. Unlike traditional feedforward neural networks, RNNs possess loops that allow information to be passed from one step in the sequence to the next, making them inherently suited for tasks where temporal dynamics and context from earlier steps are crucial.
The foundational idea behind RNNs is the concept of memory. At each step in a sequence, an RNN takes in a new input and the previous hidden state (or memory) to produce a new hidden state and an output. This recurrent nature allows RNNs to maintain a form of "memory" about previous steps in the sequence, enabling them to capture patterns over time.

% RNNs are not without challenges. They suffer from issues like the vanishing and exploding gradient problems, which make them difficult to train on long sequences. The vanishing gradient problem, in particular, means that RNNs struggle to capture long-range dependencies in data, as the influence of a given input tends to diminish over time.

% To address these challenges, more advanced RNN architectures have been developed. Long Short-Term Memory (LSTM) networks, introduced by Hochreiter and Schmidhuber in 1997, and Gated Recurrent Units (GRUs), introduced by Cho et al. in 2014, are two such architectures. Both LSTMs and GRUs incorporate gating mechanisms that regulate the flow of information, making them more effective at capturing long-range dependencies and mitigating the vanishing gradient problem. 

\begin{figure}
    \centering
    \includegraphics[width=\linewidth]{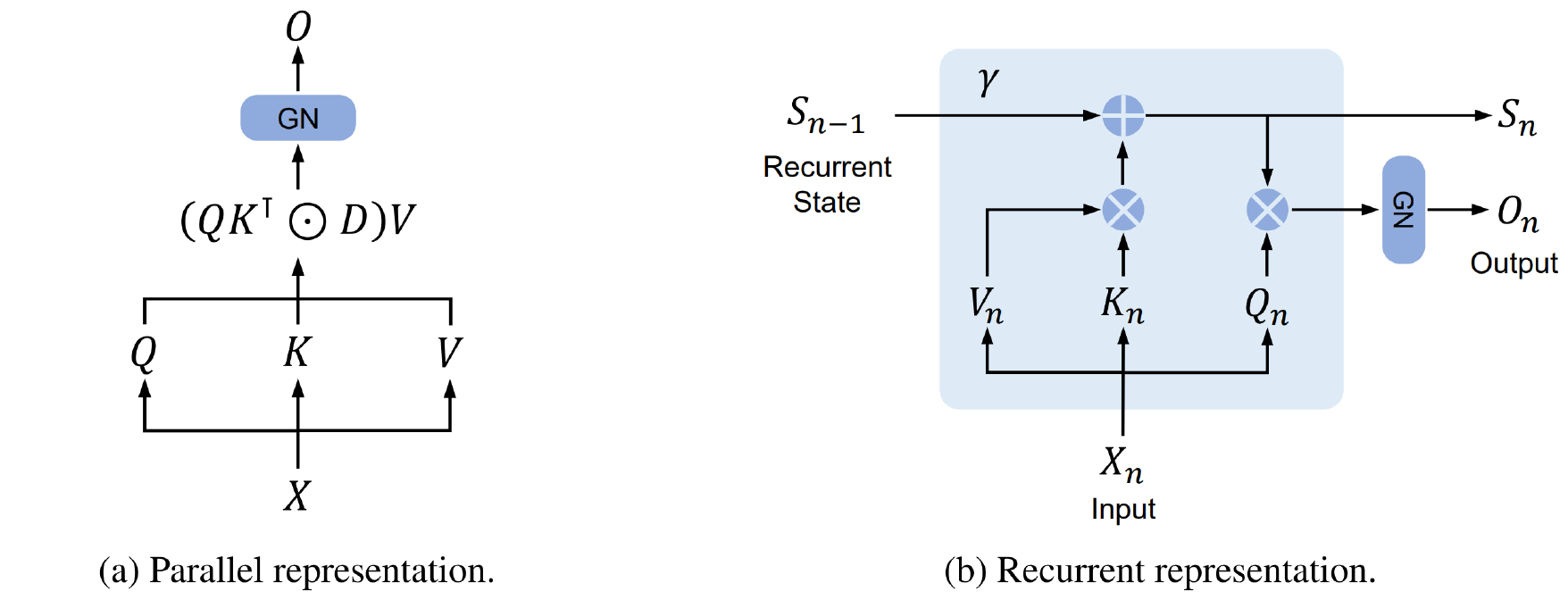}
    \caption{Dual form of RetNet. “GN” is short for GroupNorm~\cite{sun2023retentive}.}
    \label{fig:retnet}
\end{figure}

\subsection{Transformers}
\label{transformer_subsection}
Introduced in 2017 by Vaswani et al.~\cite{vaswani2017attention}, Transformers are a type of deep learning models that excel at handling sequential data, making them particularly suitable for natural language processing tasks. The key breakthrough in Transformers lies within their self-attention mechanism~\cite{zhao2020exploring}, which allows the model to assess the significance of different segments within an input sequence relative to one another. This unique capability enables the model to capture complex patterns and dependencies over long distances in the data. Unlike conventional RNNs, Transformers process input information concurrently rather than sequentially, resulting in notable enhancements in both training efficiency and scalability. Thus, it enables researchers to build large models with billions of parameters that can generate long coherent and contextual stories, answer questions accurately, and execute tasks that require logic and understanding. While they are efficient during training due to their ability to handle parallelization, one of their drawbacks is that they can be slow during inference. This is because the self-attention mechanism in Transformers requires computation over all elements of the input sequence, leading to a quadratic increase in computation as the sequence length grows. This can be particularly challenging for real-time applications or scenarios where quick response times are essential.

\subsection{RetNet}
\label{retnet_subsection}

The Retentive Network (RetNet)~\cite{sun2023retentive} emerges as a foundational architecture for large language models, offering a unique blend of training parallelism, cost-effective inference, and robust performance. At its core lies the retention mechanism tailored for sequence modeling, which seamlessly integrates three computational paradigms: parallel, recurrent, and chunkwise recurrent. This multifaceted approach ensures that the parallel representation fosters training parallelism, the recurrent representation guarantees low-cost inference, enhancing decoding throughput, latency, and GPU memory efficiency without compromising on performance. Furthermore, the chunkwise recurrent representation is adept at modeling long sequences with linear complexity, encoding each chunk in parallel but summarizing them in a recurrent manner. These distinctive features position RetNet as a potential successor to the Transformer architecture for large language models. The retention mechanism treats an 1D input sequence in a recurrent, unidirectional manner as shown in Eq~\ref{rec_eq}.

\begin{equation}
\label{rec_eq}
    o_n = \sum_{m=1}^n \gamma^{n-m}(Q_ne^{in\theta})(K_me^{im\theta})^{\dagger}v_m,
\end{equation}
where $\dagger$ is the conjugate transpose. To allow parallel training, the above equation~\ref{rec_eq} is written as follows given an input $X$:

\begin{equation}
Q = (XW_Q) \odot \Theta, \quad 
K = (XW_K) \odot \Bar{\Theta}, \quad 
V = XW_V, \quad
\Theta_n = e^{in\theta}, \quad
D_{nm} = 
\begin{cases} 
\gamma^{n-m}, & \text{if } n \geq m \\
0, & \text{if } n < m 
\end{cases}, 
\end{equation}
\begin{equation}
    \label{retention_eq}
    Retention(X) = (QK^T \odot D)V
\end{equation}

where $\Bar{\Theta}$ is the complex conjugate of $\Theta$ and $D \in \mathrm{R}^{|x|\times|x|}$ combines causal masking and exponential
decay along relative distance as one matrix. The recurrent and parallel representations of the RetNet block is shown in Fig~\ref{fig:retnet}.

\begin{figure}
    \centering
    \includegraphics[width=\linewidth]{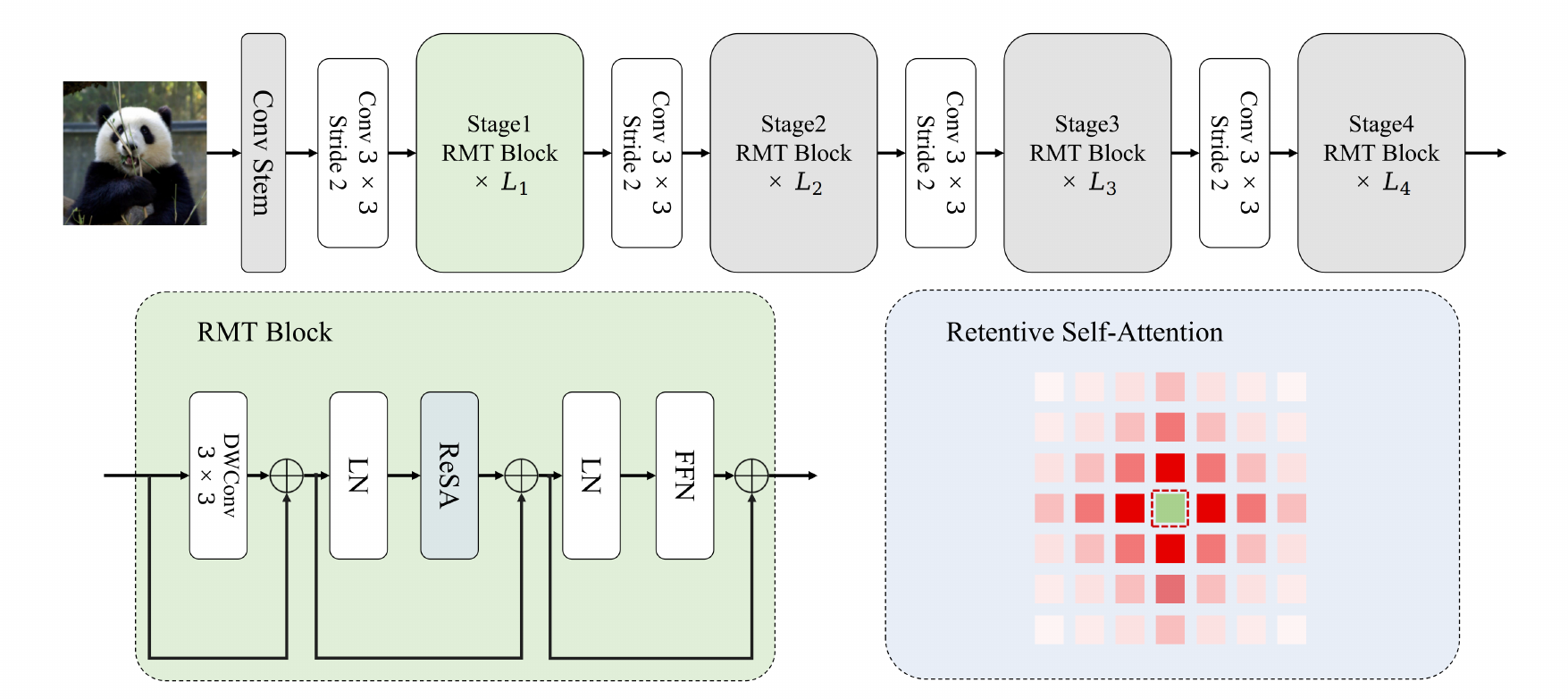}
    \caption{RetNet-based Vision Transformer~\cite{fan2023rmt}}
    \label{fig:retnet_vit}
\end{figure}

\subsection{Vision Transformers}
\label{vit_subsection}

ViTs~\cite{dosovitskiy2020image} represent a significant shift in the field of computer vision. Traditionally, Convolutional Neural Networks (CNNs)~\cite{o2015introduction} have been the dominant architecture for image-related tasks, leveraging their ability to process spatial hierarchies in images. However, the success of the Transformer architecture in natural language processing led researchers to explore its potential in the domain of vision. ViTs approach image tasks by dividing inputs into fixed-size patches, linearly embedding them, and then processing them as a sequence using the Transformer architecture. The self-attention mechanism in Transformers, which had been pivotal in capturing long-range dependencies in text, proved equally adept at handling spatial relationships in images.

\subsection{RetNet-Based Vision Transformers}
\label{retnet_vit_subsection}

As previously discussed, RetNet is mainly designed for 1D data and it treats the inputs in an unidirectional way. Efforts have been made~\cite{fan2023rmt}, as shown in Fig.~\ref{fig:retnet_vit}, to extrapolate this to the 2D space in order to be able to handle images and treat them in a bidirectional manner. The bidirectional extension of retention in Eq.~\ref{rec_eq} for each token can be written as follows:

\begin{equation}
\label{2d_retnet}
    o_n = \sum_{m=1}^N \gamma^{|n-m|}(Q_ne^{in\theta})(K_me^{im\theta})^{\dagger}v_m,
\end{equation}

where $\dagger$ is the conjugate transpose, and thus the parallel version is written as:

\begin{equation}
    \label{2d_retention_eq}
    BiRetention(X) = (QK^T \odot D^{B_i})V, \quad \text{where} \quad D^{B_i}_{nm} = \gamma^{|n-m|}
\end{equation}

Now, to extrapolate to the two-dimensional space, let's denote by $(x_n, y_n)$ the coordinate of the $n$th token of an image. The $D$ matrix is modified to contain the Manhattan distance between token pairs as follows:

\begin{equation}
\label{2d_retnet}
    D^{2d}_{nm} = \gamma^{|x_n - x_m|+|y_n - y_m|}
\end{equation}

and finally, the 2D retention can be written as:

\begin{equation}
    \label{2d_ret_eq}
    ReSA(x) = (Softmax(QK^T)\odot D^{2d})V
\end{equation}

% \section{Mathematics Behind Retentive Network}

% \subsection*{1. Parallel Representation of Retention:}
% \begin{align*}
%     Q &= (XW_Q) \odot \Theta \\
%     K &= (XW_K) \odot \Theta \\
%     V &= XW_V \\
%     \Theta_n &= e^{in\theta} \\
%     D_{nm} &= 
%     \begin{cases} 
%       \gamma^{n-m}, & \text{if } n \geq m \\
%       0, & \text{if } n < m 
%     \end{cases} \\
%     \text{Retention Output:} \quad Retention(X) &= (QK^T \odot D)V
% \end{align*}

% \subsection*{2. Recurrent Representation of Retention:}
% For the n-th timestep:
% \begin{align*}
%     S_n &= \gamma S_{n-1} + K^T_n V_n \\
%     \text{Retention Output:} \quad Retention(X_n) &= Q_n S_n
% \end{align*}

% \subsection*{3. Chunkwise Recurrent Representation of Retention:}
% For the i-th chunk:
% \begin{align*}
%     Q[i] &= QB_i : B(i+1) \\
%     K[i] &= KB_i : B(i+1) \\
%     V[i] &= VB_i : B(i+1) \\
%     R_i &= K^T[i] (V[i] \odot \zeta) + \gamma^B R_{i-1} \\
%     \text{Retention Output:} \quad Retention(X[i]) &= (Q[i] K^T[i] \odot D) V[i] + (Q[i] R_{i-1} \odot \xi)
% \end{align*}

% \subsection*{4. Gated Multi-Scale Retention:}
% \begin{align*}
%     \gamma &= 1, 2, 5 - \text{arange}(0, h) \in R^h \\
%     head_i &= Retention(X, \gamma_i) \\
%     Y &= \text{GroupNorm}_h(\text{Concat}(head_1, \dots, head_h)) \\
%     MSR(X) &= (\text{swish}(XW_G) \odot Y)W_O
% \end{align*}

\subsection{Vision-Language Models (VLMs)}
\begin{figure}
    \centering
    \includegraphics[width=\linewidth]{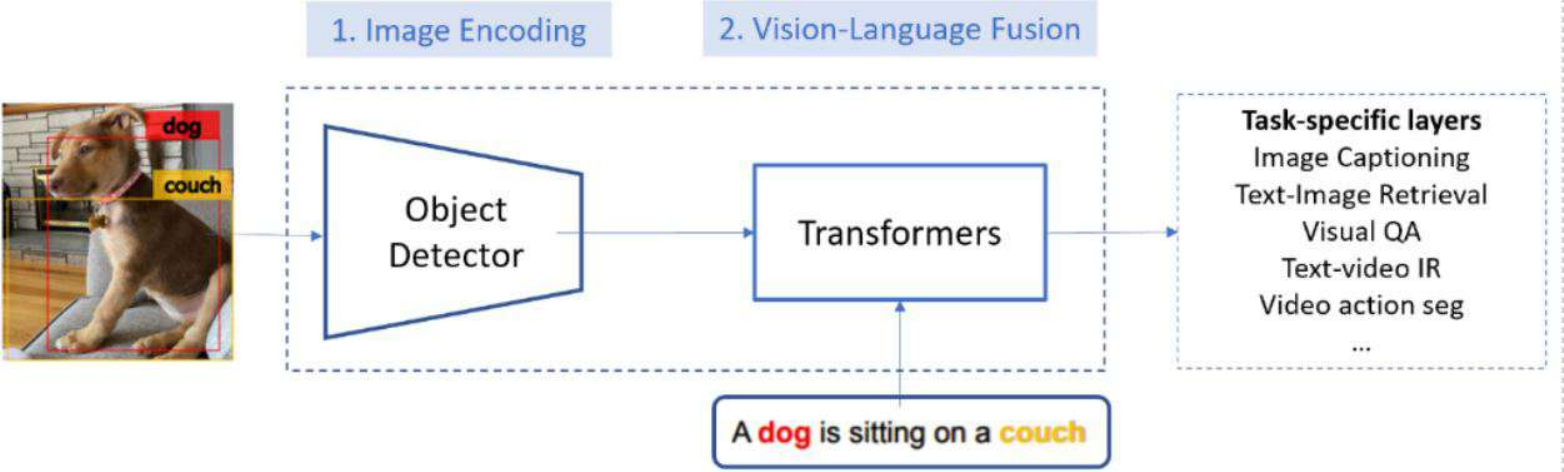}
    \caption{Vision-Language Models (VLMs)~\cite{vlm_image}}
    \label{fig:vlm}
\end{figure}

A visual language model is an AI-driven mechanism tailored to interpret and derive insights from visual content like images and videos. By training on vast collections of visual content, these models become adept at identifying and categorizing various visual components, from objects to individuals to entire scenes. VLMs have evolved significantly over the years, transitioning from the use of hand-crafted image descriptors and pre-trained word vectors to the adoption of advanced transformer architectures for both image and text encoding. Central to these models are three core components: an image encoder, a text encoder, and a fusion mechanism to integrate information from both domains as shown in Fig~\ref{fig:vlm}. The contemporary approach emphasizes strategic pre-training objectives to enhance transfer performance across various tasks. Among the notable pre-training strategies are: 
\begin{center}
\begin{itemize}
    \item \textbf{Contrastive Learning:} Aligns images and texts in a joint feature space.
    \item \textbf{PrefixLM:} Uses images as a prefix to a language model, enabling joint learning of image and text embeddings.
    \item \textbf{Multi-modal Fusing with Cross Attention:} Integrates visual data into language model layers.
    \item \textbf{MLM/ITM:} Focuses on aligning specific image parts with corresponding text using masked-language modeling and image-text matching.
    \item \textbf{No Training approach:} Leverages standalone vision and language models through iterative optimization.
\end{itemize}
\end{center}

\section{Related Work}
In this survey~\cite{zhao2023survey}, the authors presents a deep comprehensive background of LLMs, their scaling laws, and their emergent capabilities. The authors also focus on the evolution of GPT-series models and discuss their capabilities in details. Moreover, the authors discuss some available resources, datasets, available checkpoints, and APIs to help develop and train LLMs. In~\cite{hadi2023survey} the authors offer an extensive examination of LLMs, which are at the forefront of AI advancements, particularly in the realm of natural language processing. The historical progression of LLMs is charted, detailing their evolution and the various training methodologies that have been employed to enhance their capabilities. A broad spectrum of LLM applications is explored, spanning sectors such as healthcare, education, finance, and engineering, illustrating the transformative impact LLMs have on these fields. Furthermore, the paper addresses the challenges that come with implementing LLMs in practical settings. These challenges include ethical dilemmas, inherent model biases, issues with interpretability, and the substantial computational resources required for their operation. It also sheds light on strategies to improve LLM robustness and control, as well as methods to mitigate biases and enhance the quality of generated content. However, the authors of the previously mentionned surveys do not specify how these LLMs can be used in the field of computer vision. 

In~\cite{chang2023survey} the authors provide a critical examination of evaluation methods for LLMs, addressing the 'what', 'where', and 'how' of LLM assessment. It covers the range of tasks LLMs are tested on, including language processing, reasoning, and applications in various fields like medicine and education. The paper also explores the environments and benchmarks used for evaluation, and the methodologies employed to assess performance, highlighting both successes and failures of LLMs. The aim is to offer a comprehensive understanding of LLM capabilities and guide future evaluation strategies. In~\cite{long2022vision}, the authors provide a comprehensive overview of the significant strides made in the domain of Visual-Language Pretrained Models (VLPMs), which have emerged as a powerful approach for learning joint representations of visual and linguistic content. The paper begins by defining the general tasks and the overarching architecture of VLPMs, then delves into the specifics of language and vision data encoding methods. It also examines the core structures of mainstream VLPMs, detailing the nuances of their design and function. Essential strategies for both pretraining and fine-tuning these models are summarized, offering a clear roadmap for how they are developed and optimized for various tasks. However, a limitation of this paper is that it does not encompass the very latest advances in NLP, particularly the next generation of transformer models. As the field of NLP is advancing rapidly, with new models and techniques continually emerging, any overview can quickly become outdated. This paper, while thorough in its current scope, may not capture the most cutting-edge developments that have occurred since its writing, which could be crucial for researchers looking to build upon the most recent and advanced work in the field.

\section{Latest LLMs}
Large language models, are cutting-edge AI systems that can grasp and produce human-like language. Drawing inspiration from the intricacies of the human brain, they're built on intricate neural networks, such as transformer models. By learning from massive amounts of data, these models can understand context, making their text outputs feel more natural, whether they're answering your questions or spinning a story. In essence, a large language model is a sophisticated AI buddy, designed to chat and understand just like humans. While they exhibit significant potential, not all are straightforward in their implementation. LLMs, even when optimally configured, struggle with tasks necessitating precise adherence to instructions. LlamaIndex~\cite{opensourcellms} provides compatibility with a vast majority of LLMs. However, it remains ambiguous whether a specific LLM will operate efficiently immediately upon integration or if additional adjustments are imperative. The subsequent tables endeavor to assess the preliminary experience associated with diverse LlamaIndex functionalities across various LLMs. These evaluations aim to measure performance and determine the extent of modifications required to ensure optimal functionality. Typically, commercial APIs, such as OpenAI, are perceived as more dependable. Nonetheless, local open-source models are increasingly favored due to their adaptability and commitment to transparency. A comparison of various open-source and paid LLMs is shown in Table~\ref{tab:free_bllms} and Table~\ref{tab:paid_bllms}.

\begin{table}[h]
\centering
\begin{tabularx}{\textwidth}{l *{6}{p{1.9cm}}}
\toprule
Model Name & Basic\newline Query\newline Engines & Router\newline Query\newline Engine & Sub\newline Question\newline Query\newline Engine & Text2SQL & Pydantic\newline Programs & Data\newline Agents \\
\midrule
gpt-3.5-turbo  & $\checkmark$ & $\checkmark$ & $\checkmark$ & $\checkmark$ & $\checkmark$ & $\checkmark$ \\

gpt-3.5-turbo-instruct  & $\checkmark$ & $\checkmark$ & $\checkmark$ & $\checkmark$ & $\checkmark$ & \orangecircle \\

gpt-4  & $\checkmark$ & $\checkmark$ & $\checkmark$ & $\checkmark$ & $\checkmark$ & $\checkmark$ \\

claude-2 & $\checkmark$ & $\checkmark$ & $\checkmark$ & $\checkmark$ & $\checkmark$ & \orangecircle \\

claude-instant-1.2  & $\checkmark$ & $\checkmark$ & $\checkmark$ & $\checkmark$ & $\checkmark$ & \orangecircle \\

\bottomrule
\multicolumn{7}{l}{\textbf{Legend:} $\checkmark$ - Supported, \redcircle - Not Supported, \orangecircle - Partially Supported}
\end{tabularx}
\vspace{0.5cm}

\caption{Paid LLMs Benchmark~\cite{opensourcellms}}
\label{tab:paid_bllms}
\end{table}

\begin{table}[h]
\centering
\begin{tabularx}{\textwidth}{l *{6}{p{1.75cm}}}
\toprule
Model Name & Basic\newline Query\newline Engines & Router\newline Query\newline Engine & Sub\newline Question\newline Query Engine & Text2SQL & Pydantic\newline Programs & Data\newline Agents \\
\midrule
llama2-chat-7b 4bit & $\checkmark$ & \redcircle & \redcircle & \redcircle & \redcircle & \orangecircle \\
llama2-13b-chat  & $\checkmark$ & $\checkmark$ & \redcircle & $\checkmark$ & \redcircle & \redcircle\\
llama2-70b-chat  & $\checkmark$ & $\checkmark$ & \checkmark & $\checkmark$  & \redcircle & \orangecircle \\
Mistral-7B-instruct-v0.1 4bit  & $\checkmark$ & \redcircle & \redcircle & \orangecircle & \orangecircle & \orangecircle \\
zephyr-7b-alpha  & $\checkmark$ & \checkmark & \checkmark & $\checkmark$ & \checkmark & \orangecircle \\
zephyr-7b-beta  & $\checkmark$ & $\checkmark$ & \checkmark & $\checkmark$ & \redcircle & $\checkmark$ \\
\bottomrule
\multicolumn{7}{l}{\textbf{Legend:} $\checkmark$ - Supported, \redcircle - Not Supported, \orangecircle - Partially Supported}
\end{tabularx}
\vspace{0.5cm}
\caption{Open-Source LLMs Benchmark~\cite{opensourcellms}}
\label{tab:free_bllms}
\end{table}

\subsection{Applications of LLMs}

In~\cite{chen2023minigpt}, the authors introduce MiniGPT-v2, a model designed to serve as a unified interface for various vision-language tasks such as image description, visual question answering, and visual grounding. The primary challenge addressed is the use of a single model to effectively perform diverse vision-language tasks using multi-modal instructions. To achieve this, the authors propose the use of unique identifiers for different tasks during training. These identifiers help the model distinguish between task instructions and improve its learning efficiency. The model undergoes a three-stage training process. Experimental results demonstrate that MiniGPT-v2 outperforms other vision-language generalist models in various benchmarks. The architecture of MiniGPT-v2 includes a visual backbone, a linear projection layer, and a large language model. To reduce ambiguity across tasks, the authors introduce task-specific tokens in their multi-task instruction template. The model also represents spatial locations using textual formatting of bounding boxes.

Zhao et al.~\cite{Zhao_2023_CVPR}, introduces LAVILA, a new approach to learning video-language representations by leveraging Large Language Models. The paper repurpose pre-trained LLMs to be conditioned on visual input and fine-tune them to create automatic video narrators. The auto-generated narrations offer advantages such as dense coverage of long videos, better temporal synchronization of visual information and text, and higher diversity of text. The video-language embedding learned with these narrations outperforms previous state-of-the-art on multiple video tasks. LAVILA obtains significant gains in classification and retrieval benchmarks. The method shows positive scaling behavior with increasing pre-training data and model size.

Doveh et al.~\cite{doveh2023teaching}, address the challenge of enhancing VLMs' understanding of complex language structures. They introduce the concept of Structured Vision and Language Concepts (SVLC), which encompasses object attributes, relations, and states present in both text and images. The novel contribution of this work is a data-driven approach that leverages the inherent structure of language. By manipulating the textual components of existing VLM datasets, the authors emphasize SVLCs in VLM training. They generate alternative negative or positive texts, which instruct the model about SVLCs. They utilize LLMs for "unmasking" tasks, where given a sentence with a missing word, the model suggests contextually appropriate alternatives. This property is employed to create plausible negative examples, enhancing the training data for VLMs. Additionally, by prompting LLMs, the authors generate text analogies—sentences that convey similar meanings but with varied wording. This method of structured textual data manipulation, powered by LLMs, aims to emphasize the importance of SVLC in VLM training, leading to significant improvements in the model's comprehension of these concepts. Through evaluations on multiple datasets, the approach demonstrates improvements of up to 15\% in the VLMs' understanding of SVLCs, while ensuring minimal degradation in their zero-shot object recognition capabilities.

Yining et Al.~\cite{hong20233d}, made a significant push towards integrating 3D spatial understanding into LLMs. Their work  introduced a 3D-LLM framework, designed explicitly for training LLMs to comprehend and respond to 3D scenes and objects. This framework ingeniously extracts 3D features from 2D multi-view images and combines them with language prompts to generate relevant responses. One of the primary challenges addressed by the authors was the difficulty in training 3D-LLMs from scratch, primarily due to the absence of a comprehensive 3D-language dataset. To circumvent this, they proposed a method that leverages pre-trained image encoders to extract features from 2D images and map these to the 3D domain. These mapped 3D features are then seamlessly integrated into pre-trained 2D vision-language models, which subsequently serve as the foundational architecture for the 3D-LLMs. Furthermore, the study employed ChatGPT~\cite{ray2023chatgpt} to generate a diverse range of 3D-language data, emphasizing its ability to form holistic 3D descriptions from multiple image views. The results from this research underscored the potential of 3D-LLMs, showcasing their capability to generate answers closely aligned with ground-truth data and perform intricate visual reasoning tasks. This innovative approach not only bridges the existing gap between 2D image understanding and 3D spatial reasoning but also sets a promising direction for future advancements in AI-driven 3D scene comprehension.

Zhenfei et al.,~\cite{yin2023lamm} introduce an advancement in multi-modal large language models (MLLMs) by incorporating point cloud data, presenting the LAMM-Dataset and LAMM-Benchmark specifically designed for 2D image and 3D point cloud understanding. These resources are intended to support high-level vision tasks across both dimensions. They validate the effectiveness of their dataset and benchmark through extensive experiments. Furthermore, they detail the process of creating instruction-tuning datasets and benchmarks for MLLMs, aiming to expedite future research and expansion into new domains, tasks, and modalities. The paper also introduces a novel training framework for MLLMs that is optimized for the inclusion of additional modalities. Alongside this framework, they provide baseline models as well as comprehensive experimental observations and analyses to serve as a foundation for accelerating subsequent research in the field.

In this paper~\cite{wu2023pumgpt}, the authors address the challenge of product understanding in the context of online shopping, a task that involves responding to a wide range of queries based on multi-modal product information. Unlike traditional methods that use different model architectures for each sub-task, the authors introduce PUMGPT, a unified vision-language model designed to handle all aspects of product understanding within a single model structure. To effectively combine visual and textual data, they propose a novel technique called Layer-wise Adapters (LA), which aligns vision and text representations more efficiently with fewer visual tokens and allows for parameter-efficient fine-tuning. This fine-tuning capability enables PUMGPT to adapt quickly to new tasks and products. The authors also create diverse product instruction datasets using instruction templates and enhance the model's performance and generalization by incorporating open-domain datasets in the training process. PUMGPT is thoroughly evaluated and shows superior performance across a variety of product understanding tasks, including product captioning, category question-answering, attribute extraction, attribute question-answering, and free-form question-answering about products.

In this work~\cite{wang2023visionllm}, the authors tackle the limitations of vision foundation models (VFMs) in computer vision, which, despite their power, lack the open-ended task flexibility of large language models (LLMs). They introduce VisionLLM, a novel LLM-based framework designed to handle vision-centric tasks. VisionLLM treats images as if they were a foreign language, aligning vision tasks with language tasks that can be defined and manipulated through language instructions. This allows for a high degree of task customization using an LLM-based decoder that interprets these instructions to make predictions for a variety of open-ended tasks. The authors demonstrate through extensive experiments that VisionLLM can handle a range of customizations, from detailed object-level to broader task-level adjustments, with impressive results. Remarkably, the model achieves over 60\% mean Average Precision (mAP) on the COCO dataset, which is competitive with models dedicated to detection tasks. The authors aim for VisionLLM to establish a new standard for generalist models that integrate vision and language.

\section{Datasets}

A suite of datasets was introduced to enhance the training and evaluation of Large Language Models (LLMs). These datasets target various functionalities of LLMs such as API interaction, reasoning, extended context comprehension, and language understanding. A concise overview of the latest dataset is provided below:

\begin{itemize}
    \item \textbf{LAMM-Dataset~\cite{yin2023lamm}}: A diverse multi-modal dataset featuring 186K pairs of language-image instructions and responses, alongside 10K language-3D pairings, sourced from 8 image and 4 point cloud datasets. It includes four distinct types of multi-modal instruction-response pairings.

    \item \textbf{function\_calling\_extended}: A dataset consisting of English code pairs, designed to refine the API usage capabilities of LLMs. It is a human-created dataset focused on improving the interface between language models and coding queries.
    
    \item \textbf{AmericanStories~\cite{dell2023american}}: This dataset is a comprehensive corpus from the US Library of Congress for pretraining LLMs, enriching them with a wealth of American cultural and historical narratives.
    
    \item \textbf{dolma~\cite{DolmaDataset, DolmaToolkit}}: Employed for OLMo pretraining, the dolma dataset encompasses a 3 trillion token repository, aimed at providing a robust, diverse corpus for foundational language model training.
    
    \item \textbf{Platypus~\cite{platypus2023}}: Part of the Platypus2 series, this dataset with 25,000 English entries is tailored for enhancing STEM reasoning capabilities within LLMs, fostering their ability to process and reason with scientific and technical information.
    
    \item \textbf{Puffin}: The Redmond-Puffin Series is a dialog dataset with about 3,000 entries, characterized by its long conversation contexts and multi-turn dialogue structure, essential for training LLMs in sustaining coherent, context-aware interactions.
    
    \item \textbf{tiny\_series}: This dataset contains concise codes or texts in English and is devised to sharpen an LLM's problem-solving and reasoning processes through targeted, precise examples.
    
    \item \textbf{LongBench~\cite{bai2023longbench}}: Serving as an evaluative benchmark, LongBench comprises 17 tasks in both English and Chinese, testing the long-context understanding and retention in LLMs.

    \item \textbf{orca-chat~\cite{Orca-Chat}}: This dataset, consisting of 198,463 English dialog entries, is an Orca-style dialog dataset aimed at improving LLM's long-context conversational ability.
    
    \item \textbf{DialogStudio~\cite{zhang2023dialogstudio}}: A collection of diverse multilingual datasets aimed at building conversational Chatbots.
    
    \item \textbf{chatbot\_arena\_conversations~\cite{zheng2023judging}}: With 33k multilingual dialog conversations, this dataset collected cleaned conversations with pairwise human preferences on Chatbot Arena. It is used for Reinforcement Learning with Human Feedback (RLHF).
    
    \item \textbf{WebGLM-qa~\cite{liu2023webglm}}: Used by WebGLM, this English pairs dataset contains 43.6k entries. It is a QA system based on LLM and the Internet where each entry comprises a question, a response, and a reference, with the response being grounded in the reference.
    
    \item \textbf{phi-1~\cite{gunasekar2023textbooks}}: An English dialog dataset generated using the method outlined in "Textbooks Are All You Need," focusing on math and computer science problems.
    
    \item \textbf{Linly-pretraining-dataset}: A Chinese pretraining dataset used by the Linly series model, it comprises 3.4GB of data from ClueCorpusSmall, CSL news-crawl, and more.
    
    \item \textbf{FineGrainedRLHF~\cite{wu2023fine}}: Containing approximately 5,000 English examples, this repository aims to develop a new framework to collect human feedback to improve LLMs in aspects such as factual correctness and topic relevance.
    
    \item \textbf{dolphin}: With 4.5 million English pair entries, this dataset attempts to replicate Microsoft's Orca and is based on FLANv2.
    
    \item \textbf{openchat\_sharegpt4\_dataset~\cite{wang2023openchat}}: A high-quality English dialog dataset with 6k dialogs generated using GPT-4 to complete refined ShareGPT prompts, used by OpenChat.
\end{itemize}

Each dataset is uniquely curated to address specific training objectives, significantly contributing to the progression of language models in handling complex tasks and bolstering their overall proficiency. More datasets can be found in Table~\ref{tab:datasets}

\begin{center}
\begin{table}[h]
\makebox[\linewidth]{
\centering
\begin{tabularx}{\textwidth}{c c c}
\toprule
\textbf{Dataset} & \textbf{Description} & \textbf{Size} \\
\midrule
SBU~\cite{ordonez2011im2text} & Images from Flicker with captions & 1M \\
COCO Caption~\cite{chen2015microsoft} & Images from MS COCO with two versions (c5 and c40) & 330k \\
YFCC100M~\cite{thomee2016yfcc100m} & Multimedia dataset with images and videos & 100M \\
VG~\cite{krishna2017visual} & Images with object-level info, scene graphs, and etc.  & 108k \\
CC3M~\cite{sharma2018conceptual} & Image-text pairs from the web & 3.3M \\
CC12M~\cite{changpinyo2021conceptual} & Image-text pairs for VLM pretraining & 12M \\
LR~\cite{pont2020connecting} & Image captioning with local multi-modal annotations & 848,749 \\
WIT~\cite{srinivasan2021wit} & Multi-modal multilingual dataset from Wikipedia & 37.6M \\
Red Caps~\cite{desai2021redcaps} & Image-text pairs from Reddit & 12M \\
LAION400M~\cite{schuhmann2021laion} & Image-text pairs filtered by CLIP & 400M \\
LAION5B~\cite{schuhmann2022laion} & Image-text pairs in multiple languages & 5.8B \\
WuKong~\cite{gu2022wukong} & Chinese multi-modal dataset & 100M \\
CLIP~\cite{radford2021learning} & Web image-text dataset & 400M \\
ALIGN~\cite{jia2021scaling} & Noisy image-text pairs & 1.8B \\
FILIP~\cite{yao2021filip} & Image-text pairs from the internet & 300M \\
WebLI~\cite{chen2022pali} & Multilingual image-text dataset from the web & 10B\\
\bottomrule
\end{tabularx}}
\caption{Summary of Image-Text Datasets}
\label{tab:datasets}
\end{table}
\end{center}

\section{LLMs and VLMs: challnges and future research directions}
\section{LLMs}
\textbf{Designing Artificial General Intelligence (AGI) Benchmarks}: Developing benchmarks for AGI necessitates a deep exploration of the distinctions between human and machine intelligence. Human intelligence, with its nuances, adaptability, and context sensitivity, differs significantly from machine intelligence, which excels in specific tasks but often lacks adaptability and context awareness. To create comprehensive tests, these benchmarks must assess an AGI's ability to generalize knowledge across domains, reason abstractly, and learn dynamically in new, unseen scenarios. These tests should measure an AGI's adaptability, reasoning prowess, and capacity for learning on the fly. Moreover, these benchmarks should not merely aim for superhuman performance in isolated tasks but should also evaluate a balanced spectrum of abilities akin to human intelligence. This includes assessing emotional understanding, ethical decision-making, and the nuanced judgment that characterizes human-like intelligence. Crafting benchmarks that encapsulate these multi-dimensional aspects is crucial in evaluating AGI's depth, adaptability, and nuanced understanding compared to human intelligence.

\textbf{Complete Behavioral Evaluation}: Evaluating language models like LLMs solely on narrow tasks has indeed limited our understanding of their true capabilities. Shifting the focus towards holistic assessments, such as integrating LLMs with physical environments through robot control, marks a significant stride in understanding their adaptability and real-world functionality. By orchestrating a scenario where the LLM navigates and interacts in a physical setting, we delve into its capacity to process diverse streams of information—from visual and auditory cues to tactile feedback. This multifaceted evaluation tests the model's ability to synthesize various modalities seamlessly, enabling it to respond dynamically to unforeseen circumstances. It not only gauges the LLM's linguistic prowess but also sheds light on its practical application, providing a richer and more comprehensive view of its behavioral capabilities in complex, unstructured environments.

\textbf{Robustness Evaluation}: Assessing the adaptability of language models to diverse inputs like different dialects, slang, and varying grammar structures is a multifaceted challenge. First, creating comprehensive datasets that encapsulate this linguistic diversity is pivotal. These datasets must encompass a wide array of linguistic variations, capturing colloquialisms, regional nuances, and grammatical deviations. Secondly, devising evaluation metrics becomes crucial in measuring a model's robustness across these variations. Metrics should focus on consistency in understanding and generating outputs, evaluating not just accuracy but also coherence and contextual relevance. Developing these metrics requires a deep understanding of linguistic nuances and the ability to quantify these qualitative aspects objectively. Ultimately, this concerted effort aims to enhance the model's adaptability to the dynamic nature of human language, ensuring it performs reliably across diverse linguistic landscapes.

\textbf{Dynamic and Evolving Evaluation}: The rapid advancement of language models poses a challenge in evaluating their true capabilities. The static benchmarks, while useful at a point in time, tend to become outdated as models evolve. Creating evaluation protocols that dynamically adapt is crucial to ensure that these models are not merely memorizing data but are consistently learning and adapting to new challenges. By constantly presenting them with novel and unseen tasks, we can gauge their capacity to generalize knowledge, understand context, and apply reasoning skills. This dynamic evaluation approach fosters a deeper understanding of how well these models grasp the essence of language, encouraging continuous improvement and pushing the boundaries of their learning capabilities. It also enables researchers and developers to identify areas for enhancement and refinement, ensuring that these language models progress in a meaningful and sustainable manner, rather than stagnating with static benchmarks that fail to capture their evolving potential.

% \textbf{Principled and Trustworthy Evaluation}: Trust in evaluation LLMs hinges on their foundation within robust theoretical frameworks. These frameworks should be built upon established principles of fairness and lack of bias, ensuring that assessment methods are not swayed by external factors or skewed interpretations. To achieve this, evaluation tests must be meticulously designed to accurately reflect a model's performance, free from any inherent prejudice or systemic flaws. Moreover, a crucial aspect involves crafting assessments that challenge the model's capacity for generalization by generating examples beyond its training data. These out-of-distribution tests serve as litmus tests for the model's adaptability and ability to navigate novel scenarios, ultimately showcasing its true capabilities beyond the confines of familiar data patterns. The convergence of these principles—solid theoretical underpinnings, fairness, unbiased evaluation methods, and out-of-distribution testing—strengthens the reliability and trustworthiness of evaluation systems in assessing the true prowess of AI models.

\textbf{Unified Evaluation for All LLM Tasks}: LLMs indeed face a formidable challenge in catering to an extensive range of tasks, each with its distinct set of demands. One crucial aspect lies in creating an evaluation framework that can seamlessly adapt to this diversity. This framework must offer a consistent method to gauge performance, regardless of whether the task involves straightforward question answering or delves into the intricate realms of complex reasoning and creative generation. The complexity inherent in these tasks necessitates an evaluation system that not only acknowledges but effectively measures the nuances inherent in simple, direct responses as well as the intricacies demanded by abstract thinking, logical inference, and the generation of novel, imaginative content. Crafting a unified evaluation framework capable of accommodating this breadth of tasks is fundamental for comprehensively assessing the capabilities of language models, offering insights into their strengths across diverse domains and guiding their continual enhancement and refinement.

\textbf{Beyond Evaluation: LLMs Enhancement}: Evaluations are not just a final assessment but a crucial tool for continual enhancement in LLMs development. They serve as the compass guiding improvements, not merely by flagging weaknesses but by dissecting the intricacies behind robust or vulnerable responses. Understanding the reasons behind these outcomes provides a roadmap for enhancement. By delving into the 'why' behind both strengths and weaknesses, developers can pinpoint the underlying mechanisms governing language understanding. This insight is pivotal—it informs the iterative process, directing focus towards refining specific areas rather than a broad, generalized overhaul. The integration of evaluation insights into the development cycle enables a more targeted and effective approach, ultimately leading to the evolution of more proficient and nuanced Language Models.

\subsection{VLMs}
\textbf{Vision Language Interaction Modeling}: The challenge here is to create a more nuanced alignment between visual and linguistic representations. Current models often use task-level or input-level masking, which obscures parts of the data to encourage the model to learn from context. However, this does not necessarily ensure that the model is aligning the underlying features of the image with the corresponding textual features. The paper suggests that embedding-level masking, where the model learns to predict masked features directly within the embedding space, could be more effective. This approach could lead to a more granular understanding of the relationships between visual elements and their linguistic counterparts, but it requires careful consideration of how to best implement such a strategy to truly enhance the model's representational capabilities.

\textbf{VLM Pretraining Strategy}: The pretraining phase for VLMs is crucial as it sets the foundation for how well the model will perform on downstream tasks. The paper points out that there is a lack of comprehensive research on how to synergize various tasks during pretraining to benefit the model's overall performance. Multi-stage training, which involves sequentially training models on different tasks or datasets, has been explored by few but could hold the key to unlocking more effective VLMs. The challenge is multifaceted: researchers must select the right mix of datasets, design tasks that complement each other, and sequence these tasks in a way that builds upon previous learning. This process is complex and requires a deep understanding of how different tasks interact and how they can be layered to reinforce the model's learning. The ultimate goal is to develop a pretraining strategy that is tailored to enhance performance on specific tasks or domains, which would be a significant step forward for the field.

\textbf{Training Evaluation}: The current practice of evaluating VLMs primarily during downstream tasks can be inefficient and costly. If a model has fundamental flaws, these may only become apparent after significant computational resources have been expended. The paper suggests that developing intermediate metrics, akin to perplexity in language models, could provide early indicators of a model's potential performance. Such metrics would allow researchers to make adjustments during the training process, rather than after the fact, saving time and computational power. This proactive approach to evaluation could help ensure that models are on the right track before they are fully trained, but developing these metrics is challenging. They must be predictive of downstream success and sensitive enough to guide the training process effectively. This requires a deep understanding of what makes a model successful and how that success can be measured incrementally as the model learns.

\section{Conclusion}

In this survey, we effectively highlighted the significant convergence of LLMs and CV, marking a transformative phase in the realm of AI. By focusing on the evolution and advancements of transformers and their impact on ViTs and LLMs, the survey underscores the groundbreaking potential in this area. The comprehensive comparative analysis of various leading paid and open-source LLMs provides valuable insights into their performance, strengths, and areas needing improvement, while the discussion on the use of LLMs in vision-related tasks adds depth to our understanding of their practical applications. The detailed presentation of diverse datasets employed in training LLMs emphasizes the importance of varied data in enhancing the performance of these models. The survey not only serves as a rich resource for current state-of-the-art practices in the field but also opens up new avenues for future research and development. 
% \section*{Acknowledgments}
% This was was supported in full by......

%Bibliography
\bibliographystyle{unsrt}  
\bibliography{references}

\end{document}